\documentclass[11pt]{article}
\usepackage{array}
\newcolumntype{P}[1]{>{\raggedright\arraybackslash}p{#1}}

\usepackage[final]{acl}

\usepackage{times}
\usepackage{latexsym}
\usepackage{amsmath}
\usepackage{enumitem}
\usepackage{booktabs}
\usepackage{pifont}
\usepackage{xcolor}

\usepackage[T1]{fontenc}

\usepackage[utf8]{inputenc}

\usepackage{microtype}

\usepackage{inconsolata}

\usepackage{tcolorbox}
\usepackage{listings}
\usepackage{graphicx}
\usepackage{tabularx}
\usepackage{algorithm}
\usepackage{algorithmic}
\usepackage{graphicx}
\usepackage{multicol}
\usepackage{multirow}

\lstdefinestyle{plain}{
    basicstyle=\fontsize{7}{9.5}\ttfamily,
    keywordstyle=\color{blue},
    commentstyle=\color{gray},
    stringstyle=\color{green},
    showstringspaces=false,
    breaklines=true,
    breakatwhitespace=false,
    breakindent=0pt,
    escapeinside={(*@}{@*)}
}

\definecolor{MutedGreen}{RGB}{85, 170, 85}
\definecolor{CoolAccent}{RGB}{120, 145, 230}
\definecolor{IAP}{RGB}{255, 126, 121}
\definecolor{CoT}{RGB}{91, 155, 213}
\definecolor{WarmOrange}{RGB}{255, 165, 85}

\title{FinDebate: Multi-Agent Collaborative Intelligence for Financial Analysis}

\author{
  Tianshi Cai$^{1,}$\thanks{Equal contribution.}, Guanxu Li$^{1,}$\footnotemark[1], Nijia Han$^{2,}$\footnotemark[1], Ce Huang$^1$, Zimu Wang$^{1,\dag}$, \\
  \textbf{Changyu Zeng$^1$, Yuqi Wang$^4$, Jingshi Zhou$^1$, Haiyang Zhang$^1$,} \\
  \textbf{Qi Chen$^3$, Yushan Pan$^1$, Shuihua Wang$^2$, Wei Wang$^{1,}$\thanks{Corresponding authors.}} \\
  $^1$School of Advanced Technology, Xi'an Jiaotong-Liverpool University, Suzhou, China \\
  $^2$School of Science, Xi'an Jiaotong-Liverpool University, Suzhou, China \\
  $^3$School of AI and Advanced Computing, Xi'an Jiaotong-Liverpool University, Suzhou, China \\
  $^4$Shanghai Jiao Tong University, Shanghai, China \\
  \texttt{\{Tianshi.Cai24,Guanxu.Li24,Nijia.Han23\}@student.xjtlu.edu.cn} \\
  \texttt{Zimu.Wang19@student.xjtlu.edu.cn, Wei.Wang03@xjtlu.edu.cn} \\}

\begin{document}
\maketitle

\begin{abstract}
We introduce \textbf{FinDebate}, a multi-agent framework for financial analysis, integrating collaborative debate with domain-specific Retrieval-Augmented Generation (RAG). Five specialized agents, covering earnings, market, sentiment, valuation, and risk, run in parallel to synthesize evidence into multi-dimensional insights.
To mitigate overconfidence and improve reliability, we introduce a safe debate protocol that enables agents to challenge and refine initial conclusions while preserving coherent recommendations.
Experimental results, based on both LLM-based and human evaluations, demonstrate the framework's efficacy in producing high-quality analysis with calibrated confidence levels and actionable investment strategies across multiple time horizons.
\end{abstract}

\section{Introduction}

While the advent of large language models (LLMs) has catalyzed progress in NLP, the financial domain remains a high-value opportunity with strict operational and regulatory constraints, demanding accuracy, reliability, and explainability. Although LLMs can process vast volumes of unstructured financial data, their ``next token prediction,'' trained on statistical correlations, makes the outputs fluctuate across prompts and runs. As a result, confidence is often miscalibrated, and statements may appear plausible without grounding in verifiable evidence \citep{Zhang2024FinBPMAF,Tatarinov2025LanguageMF}, which are misaligned with requirements for verifiable reasoning and stable recommendations in this field.

\begin{figure*}[ht]
  \includegraphics[width=\textwidth]{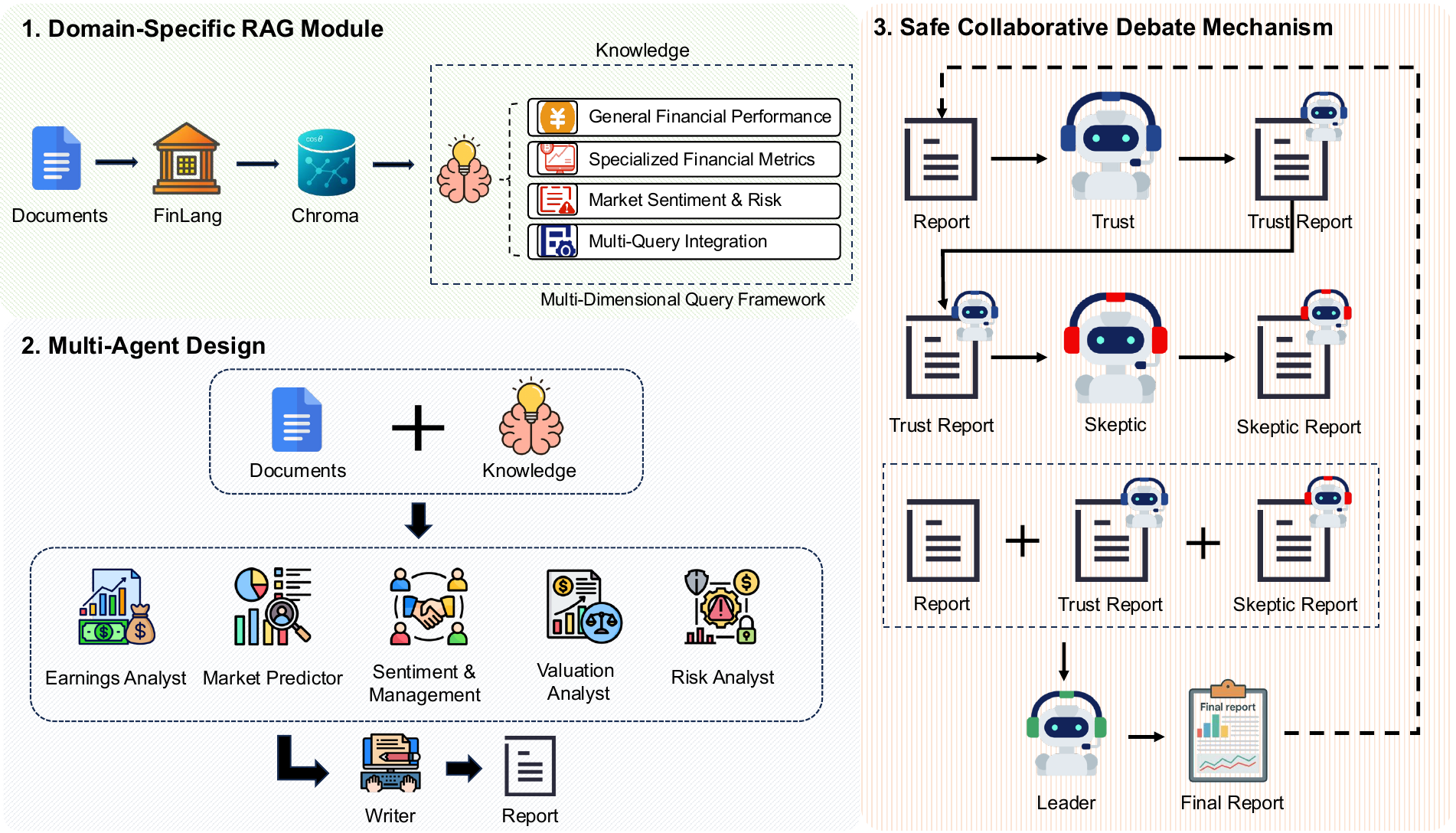}
  \caption{Overview of \textbf{FinDebate}, a multi-agent collaborative intelligence framework for financial analysis.}
  \label{fig:architecture}
  \vspace{-4mm}
\end{figure*}

Beyond the aforementioned limitations, long-form, multi-section analyst reports face document- and pipeline-level challenges.
Evidence must be synthesized into a unified, coherent narrative while avoiding topic drift and refraining from claims unsupported by the underlying transcripts \citep{Goldsack2024FromFT,Xia2025StoryWriterAM}.
A single passage can bear distinct implications across analytical dimensions, and design choices in chunking and querying materially influence what evidence is retrieved and how effectively it supports cross-aspect reasoning.
During revision cycles, stance coherence and coverage can be degraded, leading to overlooked factors and unintended shifts in the investment thesis. Moreover, reasoning must remain traceable and reference-grounded without sacrificing readability or decision-oriented clarity \cite{Li2024MEQAAB}.

Practitioners have responded with several pragmatic strategies. Template-driven workflows impose discipline and stylistic consistency but weaken the alignment between cited evidence, intervening reasoning, and the report's final stance \citep{Kang2025TemplateDrivenLF,Tian2025TemplateBasedFR}.
Retrieval-Augmented Generation (RAG) anchors factual claims, yet integrating dispersed excerpts into a cohesive, multi-faceted narrative remains challenging \citep{JimenoYepes2024FinancialRC}.
Multi-agent collaboration and debate surface issues in short-form claims, but in chaptered, long-form analyst reports, they often struggle to maintain a consistent stance while covering all essential elements \citep{Sun2024TowardsDL}. These gaps motivate an approach that jointly stabilizes stance, expands evidence coverage and explicit risk articulation, and preserves reference traceability.

To address these gaps, we introduce \textbf{FinDebate}, a safety-constrained debate protocol that stabilizes the stance while strengthening evidence and risk articulation. As shown in Figure \ref{fig:architecture}, a domain-specific RAG module and a team of role-specialized analyst agents first produce a chaptered draft. The debate phase then performs a bounded augmentation pass across roles and tasks: the pre-debate stance is fixed, roles are prohibited from changing direction, and every addition must be anchored to verifiable references. This design preserves the throughline of the investment rationale while improving coverage and verifiability, yielding analysis that remain auditable and decision-oriented, as evidenced by LLM-based and human evaluations.

\section{Methodology}
Figure \ref{fig:architecture} shows an overview of \textbf{FinDebate}, which consists of three essential modules: (1) a \textit{domain-specific RAG} module for document processing and evidence retrieval; (2) a \textit{multi-agent analysis} module for an initial draft report; (3) a \textit{debate mechanism} that yields the final report. An example of the task is shown in Appendix \ref{sec:example}. In this section, we introduce each of the modules in detail.

\subsection{Domain-Specific RAG Module}

\paragraph{Text Segmentation Strategy.}

Applying LLMs to financial analysis is constrained by limited context windows, which make it infeasible to process reports spanning hundreds of pages simultaneously. To address this, we propose a domain-specific RAG module with ChromaDB\footnote{\url{https://github.com/chroma-core/chroma}}, which enables efficient indexing and similarity search over extensive financial documents, supporting low-latency retrieval and rapid downstream processing at scale.

To mitigate the information loss by naive fixed-size chunking, we adopt a context-sensitive segmentation strategy grounded in contextual chunking \cite{günther2025latechunkingcontextualchunk}.
Instead of partitioning by fixed-length token counts, which is misaligned with the dense, highly structured nature of financial documents, we apply a recursive procedure that prioritizes semantic integrity: paragraph boundaries are preserved first, followed by sentence boundaries, and finally lexical/token boundaries.
This hierarchy prevents destructive splits, producing self-contained, interpretable segments, resulting in a robust substrate for high-precision retrieval and reliable downstream reasoning.

\paragraph{Financial Embedding and Multi-level Retrieval.}

We encode the segmented passages with FinLang\footnote{\url{https://huggingface.co/FinLang/finance-embeddings-investopedia}}, a financial embedding model adapted from BGE \cite{zhang2023retrieveaugmentlargelanguage} via domain-specific fine-tuning.
Selected for its in-domain retrieval effectiveness, FinLang captures the semantic essence of queries and grounds them in financial constructs such as investment risk, valuation metrics, market sentiment, and growth outlook.
This domain alignment enables highly precise retrieval of evidence passages, facilitating analysis of the consistency between fundamentals and stock prices and whether current valuations are justified by projected growth.

Building on multi-level retrieval \cite{adjali-etal-2024-multi}, we conduct contextual retrieval across four dimensions: \textit{general financial performance}, \textit{specialized financial metrics}, \textit{market sentiment \& risk}, and \textit{multi-query integration}  (details in Appendix \ref{sec:queries}), providing a solid analytical foundation for the subsequent multi-agent system.

\subsection{Multi-Agent Design}

Single-model approaches exhibit notable shortcomings, often yielding superficial analysis due to their reliance on generalized methodologies and limited perspectives \cite{du2023improving}.
To overcome this limitation, we propose a multi-agent collaborative framework designed to perform in-depth financial analysis across five specialized domains.
Each agent is tasked with analyzing the earnings call content from their respective domain-specific viewpoints.
Afterwards, a report synthesis module integrates these individual analysis into a unified, insightful investment advisory report.

 \subsubsection{Agent Prompting Strategy}

Each agent is equipped with a two-level prompt structure. The first level, system prompt, defines the agent's \textit{professional identity} through four key components: (1) professional credentials (e.g., a CFA charterholder with 20 years' experience), (2) an authoritative background (e.g., roles at leading investment banks and hedge funds), (3) a clear mission (e.g., to assist in institutional investment decision-making), and (4) a high-quality standard (e.g., delivering institutional-grade output).
The second level, user prompt, outlines the specific \textit{analytical task} assigned to each agent, consisting of four elements:
(1) analytical frameworks that guide systematic reasoning, (2) technical requirements specifying format and precision, (3) output specifications detailing the report structure and length, and (4) contextual integration of information retrieved through RAG. Together, these two levels ensure that the agents process both professional expertise and the ability to execute tasks effectively.

\subsubsection{Agent Specialization}

Our framework intentionally leverages five agents across different specialized analytical dimensions, establishing a holistic analytical framework that addresses the key facets of institutional investment decision-making. The core design principles of the agents are outlined below, with detailed prompts provided in Appendix \ref{sec:multiagent}:

\paragraph{\includegraphics[height=2ex]{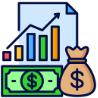} Professional Earnings Analyst} specializes in financial statement analysis and performance evaluation. Key responsibilities involve assessing revenue quality, evaluating profitability and sustainability, and examining critical financial indicators such as net interest margin (NIM), asset quality, and capital adequacy ratios.

\paragraph{\includegraphics[height=2ex]{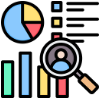} Professional Market Predictor} is tasked with forecasting market trends across multiple timeframes. This includes analyzing immediate market responses to earnings reports, evaluating the sustainability of underlying fundamental drivers, and predicting long-term market positioning based on strategic developments.

\paragraph{\includegraphics[height=2ex]{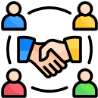} Professional Sentiment Analyst} specializes in evaluating management credibility and investor sentiment. This agent incorporates behavioral finance theories, such as anchoring effects and confirmation bias, quantifying measurable indicators like historical accuracy and transparency ratings, and translating psychological factors into actionable investment strategies.

\paragraph{\includegraphics[height=2ex]{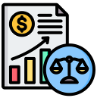} Professional Valuation Analyst} specializes in business valuation and investment recommendations. It applies a sector-specific Discounted Cashflow Model (DCF), which considers factors such as credit loss cyclicality and regulatory capital constraints, and employs dynamic weight allocation based on the reliability of various valuation methods, with a focus on verifiable business drivers.

\paragraph{\includegraphics[height=2ex]{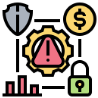} Professional Risk Analyst} provides comprehensive risk assessment and positions sizing recommendations. It evaluates various risk factors, such as credit, interest rate, and liquidity risk, while maintaining a balanced perspective to ensure realistic and actionable risk assessments.

\subsubsection{Report Synthesis}

Once the specialized agents complete their analysis, the system advances to the final stage. The Report Synthesis agent consolidates the individual outputs, extracts key financial indicators and, manages sentiment data, and generates a comprehensive report. This report is subsequently passed to the collaborative debate mechanism for further refinement, enhancing its accuracy and persuasiveness.

\subsection{Safe Collaborative Debate Mechanism}

\subsubsection{Three-Agent Collaboration}

Finally, we introduce a safe collaboration debate mechanism between three agents, motivated by established multi-agent debate methodologies \cite{du2023improving,liang2024encouraging,estornell2024multi}. It enhances the quality of the report through a single-round optimization, while maintaining the core conclusions of the original analysis. This module consists of three agents: a Trust Agent, a Skeptic Agent, and a Leader Agent, with detailed prompts provided in Appendix \ref{sec:debate}:

\paragraph{\includegraphics[height=2ex]{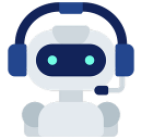} Trust Agent} enhances the original report by providing supporting evidence, reinforcing its argumentative logic, and optimizing linguistic expression. Throughout this process, it is strictly prohibited from altering the directional tone (bearish to bullish) or modifying the 1-day/1-week/1-month investment recommendations.

\paragraph{\includegraphics[height=2ex]{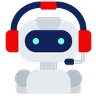} Skeptic Agent} refines the report by incorporating a risk management perspective. Its core responsibilities include identifying potential risk factors, suggesting hedge strategies, and improving the scenario analysis framework.

\paragraph{\includegraphics[height=2ex]{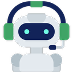} Leader Agent} synthesizes the evidence enhancements from the Trust Agent and the risk analysis from the Skeptic Agent to produce the final optimized report. The resulting content retains all core conclusions from the original report, while employing more professional and persuasive language, and offering a clearer risk-return analysis.

\subsubsection{Algorithm Design}

Algorithm \ref{alg:safe_debate} outlines the overall design of the debate framework, employing safety-first principles to preserve the integrity of the original investment recommendations. It incorporates multiple verification mechanisms while achieving systematic quality improvements through a structured optimization process.
The debate proceeds in a single round, effectively avoiding the thematic drift typically associated with multi-round iterations. Drawing from optimal rounds in related research, we compare the impacts of one-round and two-round debates, ultimately setting the maximum round to $1$. The entire process involves only minor refinements, without making directional rewrites.

It is important to note that this debate mechanism is applicable only to scenarios where reports containing pre-existing investment recommendations require further refinement. It is not intended for generating reports from scratch or for enhancing texts that lack clear directional conclusions.

\begin{algorithm}[t!]
\caption{Safe Collaborative Debate}
\label{alg:safe_debate}
\footnotesize
\begin{algorithmic}
\item[\textbf{Input:}] $R_0$ (Original\_Report), $A$ (Agent\_Analysis)
\item[\textbf{Output:}] $R^*$ (Optimized\_Report), $L$ (Debate\_Log)
\item[1:] \textbf{Safety Check:} Validate $R_0$ structure
\item[2:] \textbf{if} $\neg$has\_recommendations($R_0$) \textbf{then} return $R_0$
\item[3:] \textbf{Trust Phase:} $R_1 \leftarrow$ optimize($R_0$, $A$)
\item[4:] \quad $\circ$ Preserve core elements of $R_0$
\item[5:] \quad $\circ$ Strengthen evidence $\uparrow$
\item[6:] \textbf{Skeptic Phase:} $R_2 \leftarrow$ review($R_1$, $A$)
\item[7:] \quad $\circ$ Identify vulnerabilities in $R_1$
\item[8:] \quad $\circ$ Maintain structure integrity
\item[9:] \textbf{Leader Phase:} $R^* \leftarrow$ synthesize($R_2$, $A$)
\item[10:] \quad $\circ$ Maximize persuasive power
\item[11:] \quad $\circ$ Preserve critical elements
\item[12:] \textbf{Final Check:} Validate $R^*$ integrity
\item[13:] \textbf{if} core\_compromised($R^*$, $R_0$) \textbf{then} return $R_0$
\item[14:] \textbf{return} $R^*$, $L$
\end{algorithmic}
\end{algorithm}

\section{Experiments}

\begin{table*}[h]
\small
\centering
\begin{tabular}{lccccc}
\toprule
\textbf{Base Model} & \textbf{Zero-shot} & \textbf{Standard RAG} & \textbf{Multi-agent  w/o Debate} & \textbf{FinDebate} & \textbf{Overall Improvement} \\
\midrule
GPT-4o & $2.97 $& $3.21$ & $3.39$ & $\mathbf{3.58}$ & $\mathbf{+0.61}$ \\
Gemini 2.5 Flash & $2.90$ & $3.15$ & $3.32$ & $\mathbf{3.50}$ & $\mathbf{+0.60}$ \\
Llama 4 Maverick & $2.82$ & $3.06$ & $3.24$ & $\mathbf{3.41}$ & $\mathbf{+0.59}$ \\
DeepSeek-R1 & $2.77$ & $3.02$ & $3.10$ & $\mathbf{3.39}$ & $\mathbf{+0.62}$ \\
Claude Sonnet 4 & $3.03$ & $3.27$ & $3.45$ & $\mathbf{3.64}$ & $\mathbf{+0.61}$ \\
\bottomrule
\end{tabular}
\caption{Performance comparison of FinDebate across models. The best performance for each model is in \textbf{bold}.}
\vspace{-4mm}
\label{tab:findebate_performance}
\end{table*}

\subsection{Experimental Setup}

\paragraph{Datasets.}

We conduct experiments on the Earnings2Insights shared task \cite{takayanagi2025earnings2insights}, which focuses on generating investment guidance from earnings call transcripts. The task includes two sets of earnings call transcripts: $40$ corresponding to ECTSum \cite{mukherjee-etal-2022-ectsum}, and $24$ professional analyst reports.

\paragraph{Models and Setup.}

We employ comparative experiments using five state-of-the-art LLMs: GPT-4o (\texttt{2024-08-06}, \citealp{Hurst2024GPT4oSC}), Gemini 2.5 Flash\footnote{\url{https://deepmind.google/models/gemini/flash/}}, Llama 4 Maverick\footnote{\url{https://www.llama.com/models/llama-4/}}, DeepSeek-R1 (\texttt{0528}, \citealp{DeepSeekAI2025DeepSeekR1IR}), and Claude Sonnet 4\footnote{\url{https://www.anthropic.com/claude/sonnet/}}.
For reproducibility and a fair comparison, all models are evaluated under identical generation parameters: a temperature of $0.6$, a maximum output length of $6,500$ tokens, a top-p sampling of $0.85$, and a frequency penalty of $0.1$. Consistent prompt templates and evaluations are applied across all models.

\paragraph{Baselines.}

To demonstrate the effectiveness of FinDebate, we compare the framework against the following two baselines:
(1) \textbf{Zero-shot inference} directly processes incoming financial reports without relying on any additional information;
(2) \textbf{Standard RAG} represents the traditional RAG approach with a general-purpose embedding model;
(3) \textbf{Multi-agent generation} serves as an ablation study that removes the safe collaborative debate mechanism, so as to assess the contribution of the debate mechanism itself.

\paragraph{Evaluation Metrics.}

To ensure a rigorous yet tractable evaluation process, we sample $10$ reports from the ECTSum dataset and $5$ from the new, professional subset, assessing the quality of the models' financial analysis. Following the framework of \newcite{Goldsack2024FromFT}, we define an evaluation protocol spanning two core dimensions and implement it using GPT-4o \cite{Hurst2024GPT4oSC}: (1) Textual Quality, covering \textit{readability}, \textit{linguistic abstractness}, and \textit{coherence}; and (2) Financial Analysis Professionalism, encompassing \textit{financial key point coverage}, \textit{background context adequacy}, \textit{management sentiment conveyance}, \textit{future outlook analysis}, and \textit{factual accuracy}.
Each report is on a four-point scale ($1$ = poor to $4$ = excellent). Detailed definitions of these dimensions and an illustrative prompt are provided in Appendix \ref{sec:eval}.

Human evaluation is also conducted, with a primary focus on whether the report can effectively guide and persuade investors to make correct decisions, including the average accuracy of investment choices (\textit{Long} or \textit{Short}) made by experts for the next day, week, and month based on the reports, and the average Likert scores on clarity, logic, persuasiveness, readability, and usefulness.

\subsection{Results and Analysis}

\begin{figure}[t!]
    \centering
    \includegraphics[width=\linewidth]{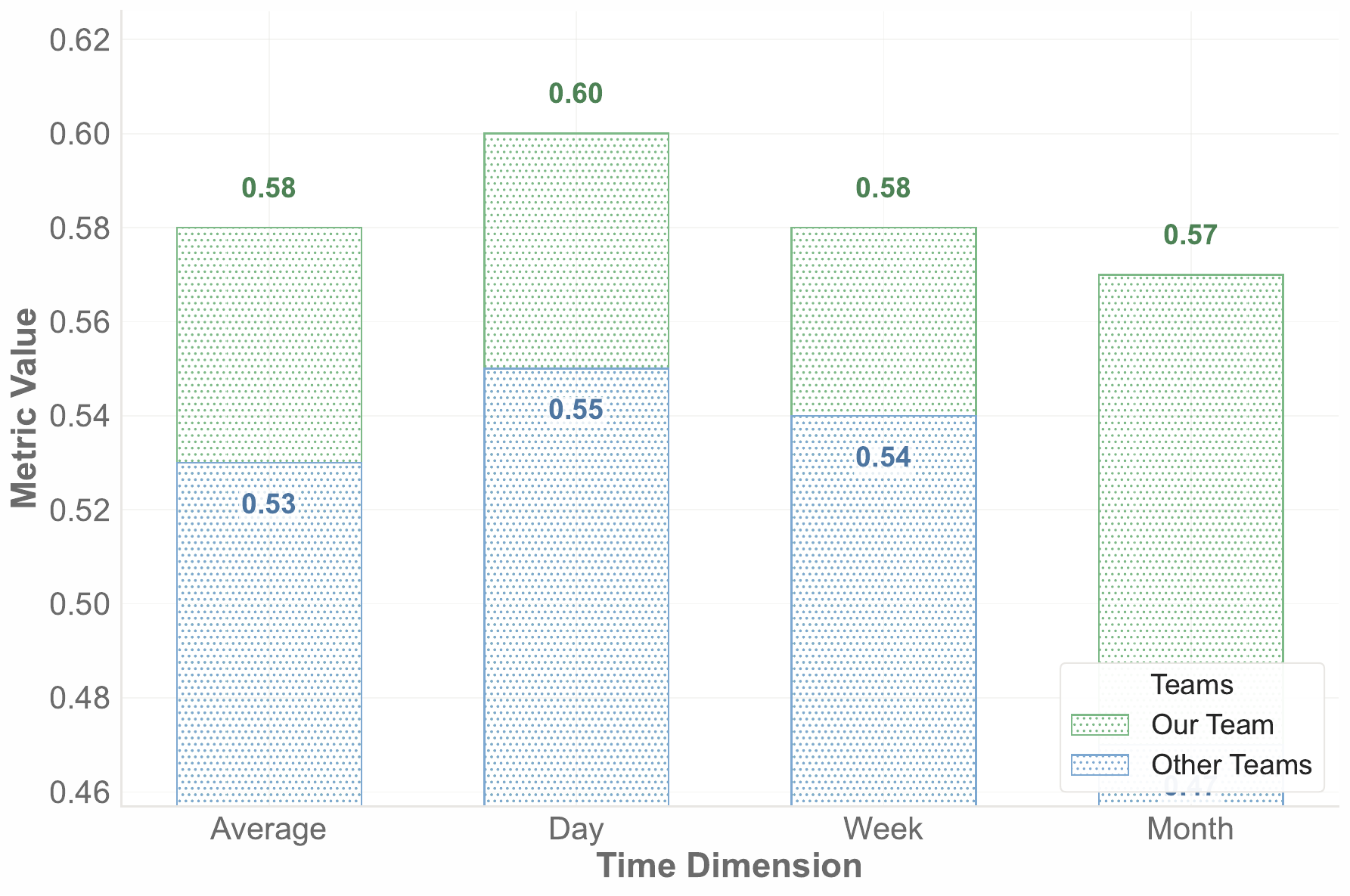}
    \caption{Human evaluation results on financial decision accuracy.}
    \vspace{-4mm}
    \label{fig:decision}
\end{figure}

\begin{figure}[t!]
    \centering
    \includegraphics[width=\linewidth]{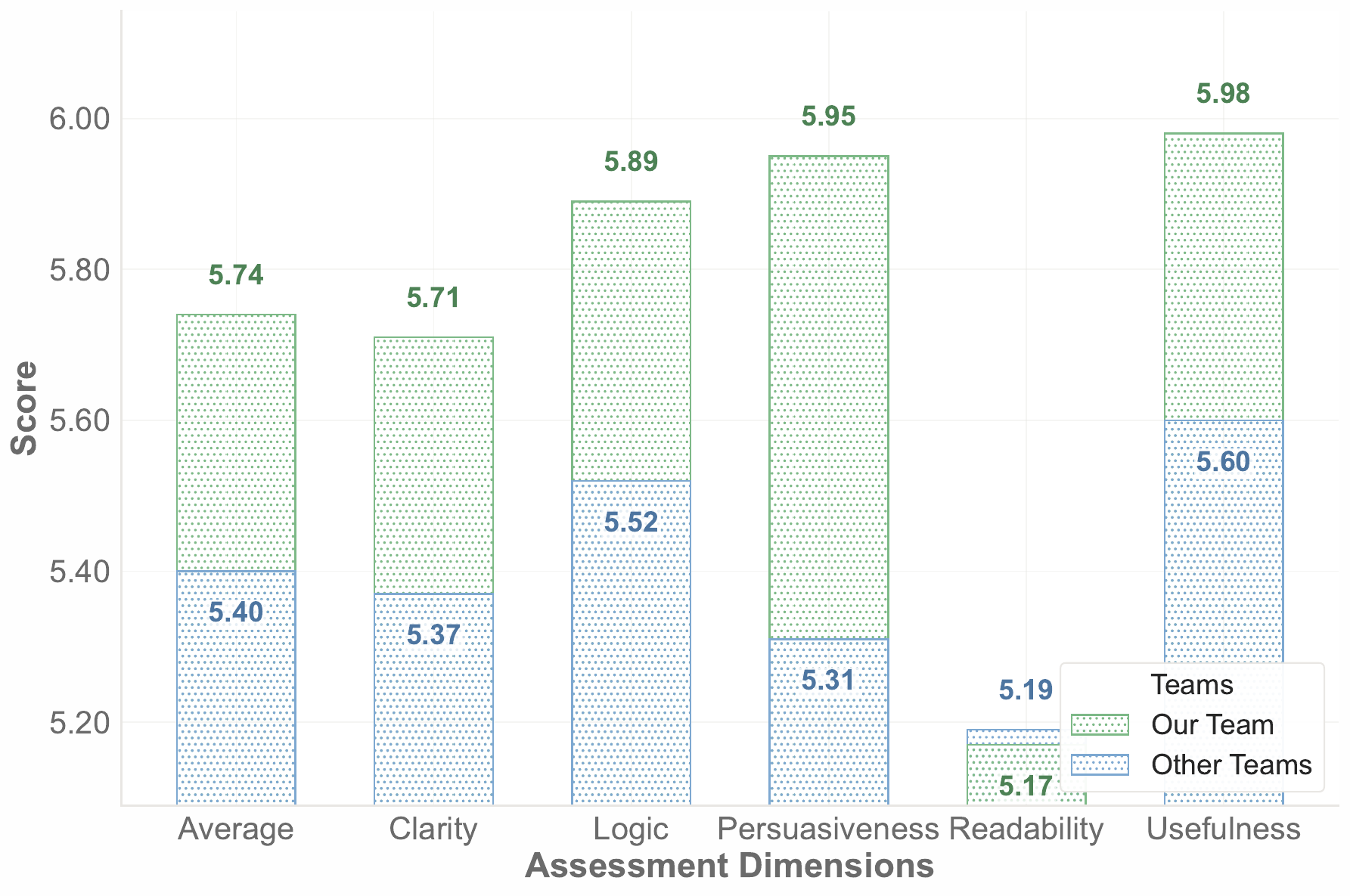}
    \caption{Human evaluation results on financial report quality.}
    \vspace{-4mm}
    \label{fig:quality}
\end{figure}

Table \ref{tab:findebate_performance} presents the main results of FinDebate compared to the zero-shot inference, standard RAG, and multi-agent generation baselines, and Figures 2 and 3 visualize human evaluation results. In comparison to practitioners, FinDebate demonstrates substantial improvements in financial decision prediction, achieving superior performance in clarity, logic, persuasiveness, and usefulness.

Our FinDebate framework consistently delivers substantial improvements across all five models, with performance gains ranging from $0.59$ to $0.62$, resulting in an average enhancement of $20.4\%$. These improvements are statistically significant ($p < 0.001$, via paired $t$-tests).
FinDebate elevates performance from ``satisfactory'' levels ($\sim3.0$ points) to ``excellent'' standards ($\sim3.6$ points), highlighting the distinctive value of collaborative intelligence in complex reasoning tasks.
This cross-model consistency further emphasizes the framework’s universality and technical superiority. 
By transforming AI-driven financial analysis from a tool-assisted approach to a professional analyst-level capability, FinDebate establishes a foundation for real-world applications through its model-agnostic design and structured collaborative methodology.

\section{Conclusion and Future Work}

We introduce FinDebate, a multi-agent framework that integrates domain-specific RAG, specialized analytical agents, and a safe collaborative debate mechanism for financial analysis, generating institutional-grade financial reports with actionable, multi-horizon investment recommendations, effectively addressing key limitations in existing financial AI applications.
In the future, we will extend this framework to broader financial domains, developing dynamic confidence adjustment mechanisms, and integrating with real-time market data. We will also transfer this system to other applications.

\bibliography{custom}

\begin{thebibliography}{19}
\providecommand{\natexlab}[1]{#1}

\bibitem[{Adjali et~al.(2024)Adjali, Ferret, Ghannay, and Borgne}]{adjali-etal-2024-multi}
Omar Adjali, Olivier Ferret, Sahar Ghannay, and Herv{\'e}~Le Borgne. 2024.
\newblock \href {https://api.semanticscholar.org/CorpusID:273901161} {Multi-level information retrieval augmented generation for knowledge-based visual question answering}.
\newblock In \emph{Conference on Empirical Methods in Natural Language Processing}.

\bibitem[{Du et~al.(2023)Du, Li, Torralba, Tenenbaum, and Mordatch}]{du2023improving}
Yilun Du, Shuang Li, Antonio Torralba, Joshua~B. Tenenbaum, and Igor Mordatch. 2023.
\newblock \href {https://api.semanticscholar.org/CorpusID:258841118} {Improving factuality and reasoning in language models through multiagent debate}.
\newblock \emph{ArXiv}, abs/2305.14325.

\bibitem[{Estornell and Liu(2024)}]{estornell2024multi}
Andrew Estornell and Yang Liu. 2024.
\newblock \href {https://api.semanticscholar.org/CorpusID:276185075} {Multi-llm debate: Framework, principals, and interventions}.
\newblock In \emph{Neural Information Processing Systems}.

\bibitem[{Goldsack et~al.(2024)Goldsack, Wang, Lin, and Chen}]{Goldsack2024FromFT}
Tomas Goldsack, Yang Wang, Chen Lin, and Chung-Chi Chen. 2024.
\newblock \href {https://api.semanticscholar.org/CorpusID:273025963} {From facts to insights: A study on the generation and evaluation of analytical reports for deciphering earnings calls}.
\newblock In \emph{International Conference on Computational Linguistics}.

\bibitem[{Gunther et~al.(2024)Gunther, Mohr, Wang, and Xiao}]{günther2025latechunkingcontextualchunk}
Michael Gunther, Isabelle Mohr, Bo~Wang, and Han Xiao. 2024.
\newblock \href {https://api.semanticscholar.org/CorpusID:272524899} {Late chunking: Contextual chunk embeddings using long-context embedding models}.
\newblock \emph{ArXiv}, abs/2409.04701.

\bibitem[{Guo et~al.(2025)Guo, Yang, Zhang, Song, Zhang, Xu, Zhu, Ma, Wang, Bi, Zhang, Yu, Wu, Wu, Gou, Shao, Li, Gao, Liu, Xue, Wang, Wu, Feng, Lu, Zhao, Deng, Zhang, Ruan, Dai, Chen, Ji, Li, Lin, Dai, Luo, Hao, Chen, Li, Zhang, Bao, Xu, Wang, Ding, Xin, Gao, Qu, Li, Guo, Li, Wang, Chen, Yuan, Qiu, Li, Cai, Ni, Liang, Chen, Dong, Hu, Gao, Guan, Huang, Yu, Wang, Zhang, Zhao, Wang, Zhang, Xu, Xia, Zhang, Zhang, Tang, Li, Wang, Li, Tian, Huang, Zhang, Wang, Chen, Du, Ge, Zhang, Pan, Wang, Chen, Jin, Chen, Lu, Zhou, Chen, Ye, Wang, Yu, Zhou, Pan, Li, Zhou, Wu, Yun, Pei, Sun, Wang, Zeng, Zhao, Liu, Liang, Gao, Yu, Zhang, Xiao, An, Liu, Wang, aokang Chen, Nie, Cheng, Liu, Xie, Liu, Yang, Li, Su, Lin, Li, Jin, Shen, Chen, Sun, Wang, Song, Zhou, Wang, Shan, Li, Wang, Wei, Zhang, Xu, Li, Zhao, Sun, Wang, Yu, Zhang, Shi, Xiong, He, Piao, Wang, Tan, Ma, Liu, Guo, Ou, Wang, Gong, Zou, He, Xiong, Luo, mei You, Liu, Zhou, Zhu, Huang, Li, Zheng, Zhu, Ma, Tang, Zha, Yan, Ren, Ren, Sha, Fu, Xu, Xie, guo Zhang, Hao, Ma, Yan,
  Wu, Gu, Zhu, Liu, Li, Xie, Song, Pan, Huang, Xu, Zhang, and Zhang}]{DeepSeekAI2025DeepSeekR1IR}
Daya Guo, Dejian Yang, Haowei Zhang, Jun-Mei Song, Ruoyu Zhang, Runxin Xu, Qihao Zhu, Shirong Ma, Peiyi Wang, Xiaoling Bi, Xiaokang Zhang, Xingkai Yu, Yu~Wu, Z.~F. Wu, Zhibin Gou, Zhihong Shao, Zhuoshu Li, Ziyi Gao, Aixin Liu, and 178 others. 2025.
\newblock \href {https://api.semanticscholar.org/CorpusID:275789950} {Deepseek-r1: Incentivizing reasoning capability in llms via reinforcement learning}.
\newblock \emph{ArXiv}, abs/2501.12948.

\bibitem[{Hurst et~al.(2024)Hurst, Lerer, Goucher, Perelman, Ramesh, Clark, Ostrow, Welihinda, Hayes, Radford, Mkadry, Baker-Whitcomb, Beutel, Borzunov, Carney, Chow, Kirillov, Nichol, Paino, Renzin, Passos, Kirillov, Christakis, Conneau, Kamali, Jabri, Moyer, Tam, Crookes, Tootoochian, Tootoonchian, Kumar, Vallone, Karpathy, Braunstein, Cann, Codispoti, Galu, Kondrich, Tulloch, drey Mishchenko, Baek, Jiang, toine Pelisse, Woodford, Gosalia, Dhar, Pantuliano, Nayak, Oliver, Zoph, Ghorbani, Leimberger, Rossen, Sokolowsky, Wang, Zweig, Hoover, Samic, McGrew, Spero, Giertler, Cheng, Lightcap, Walkin, Quinn, Guarraci, Hsu, Kellogg, Eastman, Lugaresi, Wainwright, Bassin, Hudson, Chu, Nelson, Li, Shern, Conger, Barette, Voss, Ding, Lu, Zhang, Beaumont, Hallacy, Koch, Gibson, Kim, Choi, McLeavey, Hesse, Fischer, Winter, Czarnecki, Jarvis, Wei, Koumouzelis, Sherburn, Kappler, Levin, Levy, Carr, Farhi, M{\'e}ly, Robinson, Sasaki, Jin, Valladares, Tsipras, Li, Nguyen, Findlay, Oiwoh, Wong, Asdar, Proehl, Yang, Antonow,
  Kramer, Peterson, Sigler, Wallace, Brevdo, Mays, Khorasani, Such, Raso, Zhang, von Lohmann, Sulit, Goh, Oden, Salmon, Starace, Brockman, Salman, Bao, Hu, Wong, Wang, Schmidt, Whitney, woo Jun, Kirchner, de~Oliveira~Pinto, Ren, Chang, Chung, Kivlichan, O’Connell, Osband, Silber, Sohl, Okuyucu, Lan, Kostrikov, Sutskever, Kanitscheider, Gulrajani, Coxon, Menick, Pachocki, Aung, Betker, Crooks, Lennon, Kiros, Leike, Park, Kwon, Phang, Teplitz, Wei, Wolfe, Chen, Harris, Varavva, Lee, Shieh, Lin, Yu, Weng, Tang, Yu, Jang, Candela, Beutler, Landers, Parish, Heidecke, Schulman, Lachman, McKay, Uesato, Ward, Kim, Huizinga, Sitkin, Kraaijeveld, Gross, Kaplan, Snyder, Achiam, Jiao, Lee, Zhuang, Harriman, Fricke, Hayashi, Singhal, Shi, Karthik, Wood, Rimbach, Hsu, Nguyen, Gu-Lemberg, Button, Liu, Howe, Muthukumar, Luther, Ahmad, Kai, Itow, Workman, Pathak, Chen, Jing, Guy, Fedus, Zhou, Mamitsuka, Weng, McCallum, Held, Long, Feuvrier, Zhang, Kondraciuk, Kaiser, Hewitt, Metz, Doshi, Aflak, Simens, laine Boyd, Thompson,
  Dukhan, Chen, Gray, Hudnall, Zhang, Aljubeh, teusz Litwin, Zeng, Johnson, Shetty, Gupta, Shah, Yatbaz, Yang, Zhong, Glaese, Chen, Janner, Lampe, Petrov, Wu, Wang, Fradin, Pokrass, Castro, Castro, Pavlov, Brundage, Wang, Khan, Murati, Bavarian, Lin, Yesildal, Soto, Gimelshein, talie Cone, Staudacher, Summers, LaFontaine, Chowdhury, Ryder, Stathas, Turley, Tezak, Felix, Kudige, Keskar, Deutsch, Bundick, Puckett, Nachum, Okelola, Boiko, Murk, Jaffe, Watkins, Godement, Campbell-Moore, Chao, McMillan, Belov, Su, Bak, Bakkum, Deng, Dolan, Hoeschele, Welinder, Tillet, Pronin, Tillet, Dhariwal, ing Yuan, Dias, Lim, Arora, Troll, Lin, Lopes, Puri, Miyara, Leike, Gaubert, Zamani, Wang, Donnelly, Honsby, Smith, Sahai, Ramchandani, Huet, Carmichael, Zellers, Chen, Chen, Nigmatullin, Cheu, Jain, Altman, Schoenholz, Toizer, Miserendino, Agarwal, Culver, Ethersmith, Gray, Grove, Metzger, Hermani, Jain, Zhao, Wu, Jomoto, Wu, Xia, Phene, Papay, Narayanan, Coffey, Lee, Hall, Balaji, Broda, Stramer, Xu, Gogineni,
  Christianson, Sanders, Patwardhan, Cunninghman, Degry, Dimson, Raoux, Shadwell, Zheng, Underwood, Markov, Sherbakov, Rubin, Stasi, Kaftan, Heywood, Peterson, Walters, Eloundou, Qi, Moeller, Monaco, Kuo, Fomenko, Chang, Zheng, Zhou, Manassra, Sheu, Zaremba, Patil, Qian, Kim, Cheng, Zhang, He, Zhang, Jin, Dai, and Malkov}]{Hurst2024GPT4oSC}
Aaron Hurst, Adam Lerer, Adam~P. Goucher, Adam Perelman, Aditya Ramesh, Aidan Clark, AJ~Ostrow, Akila Welihinda, Alan Hayes, Alec Radford, Aleksander Mkadry, Alex Baker-Whitcomb, Alex Beutel, Alex Borzunov, Alex Carney, Alex Chow, Alexander Kirillov, Alex Nichol, Alex Paino, and 397 others. 2024.
\newblock \href {https://api.semanticscholar.org/CorpusID:273662196} {Gpt-4o system card}.
\newblock \emph{ArXiv}, abs/2410.21276.

\bibitem[{Jimeno-Yepes et~al.(2024)Jimeno-Yepes, You, Milczek, Laverde, and Li}]{JimenoYepes2024FinancialRC}
Antonio Jimeno-Yepes, Yao You, Jan Milczek, Sebastian Laverde, and Ren-Yu Li. 2024.
\newblock \href {https://api.semanticscholar.org/CorpusID:267547721} {Financial report chunking for effective retrieval augmented generation}.
\newblock \emph{ArXiv}, abs/2402.05131.

\bibitem[{Kang et~al.(2025)Kang, Wang, Jin, Wang, Huang, and Wang}]{Kang2025TemplateDrivenLF}
Xiaoqiang Kang, Zimu Wang, Xiao-Bo Jin, Wei Wang, Kaizhu Huang, and Qiufeng Wang. 2025.
\newblock \href {https://api.semanticscholar.org/CorpusID:277743706} {Template-driven llm-paraphrased framework for tabular math word problem generation}.
\newblock In \emph{AAAI Conference on Artificial Intelligence}.

\bibitem[{Li et~al.(2024)Li, Wang, Tran, Xia, and Du}]{Li2024MEQAAB}
Ruosen Li, Zimu Wang, Son~Quoc Tran, Lei Xia, and Xinya Du. 2024.
\newblock \href {https://api.semanticscholar.org/CorpusID:276318394} {Meqa: A benchmark for multi-hop event-centric question answering with explanations}.
\newblock In \emph{Neural Information Processing Systems}.

\bibitem[{Liang et~al.(2024)Liang, He, Jiao, Wang, Wang, Wang, Yang, Tu, and Shi}]{liang2024encouraging}
Tian Liang, Zhiwei He, Wenxiang Jiao, Xing Wang, Yan Wang, Rui Wang, Yujiu Yang, Zhaopeng Tu, and Shuming Shi. 2024.
\newblock \href {https://api.semanticscholar.org/CorpusID:258967540} {Encouraging divergent thinking in large language models through multi-agent debate}.
\newblock In \emph{Conference on Empirical Methods in Natural Language Processing}.

\bibitem[{Mukherjee et~al.(2022)Mukherjee, Bohra, Banerjee, Sharma, Hegde, Shaikh, Shrivastava, Dasgupta, Ganguly, Ghosh, and Goyal}]{mukherjee-etal-2022-ectsum}
Rajdeep Mukherjee, Abhinav Bohra, Akash Banerjee, Soumya Sharma, Manjunath Hegde, Afreen Shaikh, Shivani Shrivastava, K.~Dasgupta, Niloy Ganguly, Saptarshi Ghosh, and Pawan Goyal. 2022.
\newblock \href {https://api.semanticscholar.org/CorpusID:253098682} {Ectsum: A new benchmark dataset for bullet point summarization of long earnings call transcripts}.
\newblock In \emph{Conference on Empirical Methods in Natural Language Processing}.

\bibitem[{Sun et~al.(2024)Sun, Li, Zhong, Zhao, and Yan}]{Sun2024TowardsDL}
Xiaoxi Sun, Jinpeng Li, Yan Zhong, Dongyan Zhao, and Rui Yan. 2024.
\newblock \href {https://api.semanticscholar.org/CorpusID:270258419} {Towards detecting llms hallucination via markov chain-based multi-agent debate framework}.
\newblock In \emph{IEEE International Conference on Acoustics, Speech, and Signal Processing}.

\bibitem[{Takayanagi et~al.(2025)Takayanagi, Goldsack, Izumi, Lin, Takamura, and Chen}]{takayanagi2025earnings2insights}
Takehiro Takayanagi, Tomas Goldsack, Kiyoshi Izumi, Chenghua Lin, Hiroya Takamura, and Chung-Chi Chen. 2025.
\newblock \href {<https://sigfintech.github.io/fineval.html>} {{Earnings2Insights: Analyst Report Generation for Investment Guidance}}.
\newblock In \emph{Proceedings of the FinNLP Workshop at EMNLP 2025}, Suzhou, China.
\newblock Overview paper for the Earnings2Insights shared task (FinEval) at FinNLP 2025.

\bibitem[{Tatarinov et~al.(2025)Tatarinov, Sukhani, Shah, and Chava}]{Tatarinov2025LanguageMF}
Nikita Tatarinov, Siddhant Sukhani, Agam Shah, and Sudheer Chava. 2025.
\newblock \href {https://api.semanticscholar.org/CorpusID:277667670} {Language modeling for the future of finance: A quantitative survey into metrics, tasks, and data opportunities}.
\newblock \emph{ArXiv}, abs/2504.07274.

\bibitem[{Tian et~al.(2025)Tian, Tang, Wang, Yen, and Peng}]{Tian2025TemplateBasedFR}
Yong-En Tian, Yu-Chien Tang, Kuang-Da Wang, An-Zi Yen, and Wen-Chih Peng. 2025.
\newblock \href {https://api.semanticscholar.org/CorpusID:277955491} {Template-based financial report generation in agentic and decomposed information retrieval}.
\newblock \emph{ArXiv}, abs/2504.14233.

\bibitem[{Xia et~al.(2025)Xia, Peng, Qi, Wang, Xu, Hou, and Li}]{Xia2025StoryWriterAM}
Haotian Xia, Hao Peng, Yunjia Qi, Xiaozhi Wang, Bin Xu, Lei Hou, and Juanzi Li. 2025.
\newblock \href {https://api.semanticscholar.org/CorpusID:279464402} {Storywriter: A multi-agent framework for long story generation}.
\newblock \emph{ArXiv}, abs/2506.16445.

\bibitem[{Zhang et~al.(2023)Zhang, Xiao, Liu, Dou, and Nie}]{zhang2023retrieveaugmentlargelanguage}
Peitian Zhang, Shitao Xiao, Zheng Liu, Zhicheng Dou, and Jian-Yun Nie. 2023.
\newblock \href {https://api.semanticscholar.org/CorpusID:263835099} {Retrieve anything to augment large language models}.
\newblock \emph{ArXiv}, abs/2310.07554.

\bibitem[{Zhang et~al.(2024)Zhang, Sen, Wang, Sun, Jiang, and Su}]{Zhang2024FinBPMAF}
Zhilu Zhang, Procheta Sen, Zimu Wang, Ruoyu Sun, Zhengyong Jiang, and Jionglong Su. 2024.
\newblock \href {https://api.semanticscholar.org/CorpusID:268417098} {Finbpm: A framework for portfolio management-based financial investor behavior perception model}.
\newblock In \emph{Conference of the European Chapter of the Association for Computational Linguistics}.

\end{thebibliography}

\appendix
\onecolumn

\section{Dataset Example}
\label{sec:example}

\begin{figure*}[h!]
    \begin{tcolorbox}[title=Input Financial Earnings Call of the Example, left=2mm,right=1mm,top=-3mm, bottom=0mm,colback=white,colframe=MutedGreen]
    \begin{lstlisting}[style=plain]

## Financial Earnings Call

### Prepared remarks
**Operator**
: Greetings, and welcome to the ABM Industries Incorporated Third Quarter 2021 Earnings Call. [Operator Instructions] As a reminder, this conference is being recorded. It is now my pleasure to introduce David Gold, Investor and Media Relations. Thank you, you may begin.

**Investor Relations**
: Thank you for joining us this morning. With us today are Scott Salmirs, our President and Chief Executive Officer; and Earl Ellis, our Executive Vice President and Chief Financial Officer. We issued our press release yesterday afternoon announcing our third quarter fiscal 2021 financial results. A copy of this release and an accompanying slide presentation can be found on our corporate website. Before we begin, I would like to remind you that our call and presentation today contain predictions, estimates and other forward-looking statements. Our use of the words estimate, expect, and similar expressions are intended to identify these statements. Statements represent our current judgment of what the future holds. While we believe them to be reasonable, these statements are subject to risks and uncertainties that could cause our actual results to differ materially. These factors are described in a slide that accompanies our presentation, as well as our filings with the SEC. During the course of this call, certain non-GAAP financial information will be presented. A reconciliation of historical non-GAAP numbers to GAAP financial measures is available at the end of the presentation and on the company's website under the Investor tab. I would now like to turn the call over to Scott.

**CEO**
: Thanks, David. Good morning, and thank you all for joining us today to discuss our third quarter results. As detailed in yesterday's release, ABM generated strong third quarter results featuring double-digit growth in revenue, continued solid cash generation, and a 20% gain in adjusted earnings per share. Revenue growth was broad-based as each of our five business segments achieved year-over-year gains in revenue, aided by an improving business environment and the gradual reopening of the economy. Our team members once again executed well and continue to provide exceptional service to our clients. Overall, demand for ABM's higher margin virus protection services remained elevated in the quarter, underscoring ongoing client concerns regarding cleaning and disinfection of their facilities. As anticipated, demand for virus protection eased slightly in the third quarter compared to the second quarter of fiscal 2021, but remain well above pre-pandemic levels. The emergence of the Delta variant and rising COVID-19 cases nationally have gains heightened interest in the need for disinfection prevention measures, particularly in high traffic areas. As we look forward to 2022 and beyond, we believe that virus protection services will remain a contributor to our overall revenue as disinfection becomes a standard service protocol and facility maintenance programs. [...]
    \end{lstlisting}
    \end{tcolorbox}

    \begin{tcolorbox}[title=Output Financial Analysis of the Example, left=2mm,right=1mm,top=-3mm, bottom=0mm,colback=white,colframe=MutedGreen]
    \begin{lstlisting}[style=plain]

- abm industries q3 adjusted earnings per share $0.90.

- q3 gaap loss per share $0.20 from continuing operations.

- q3 adjusted earnings per share $0.90.

- raises adjusted earnings per share guidance for full year fiscal 2021.

- q3 revenue rose 10.7 percent to $1.54 billion.

- increasing guidance for full year 2021 adjusted income from continuing operations to $3.45 to $3.55 per share.
    \end{lstlisting}
    \end{tcolorbox}
    \caption{Dataset example of the Earnings2Insights shared task \cite{takayanagi2025earnings2insights}, where the example is from the ECTSum subset \cite{mukherjee-etal-2022-ectsum}. Models receive an input financial earnings call with management remarks, Q\&A sessions, etc., and generate a structured financial analysis report for investment recommendation.}
    \label{fig:data example}
\end{figure*}

\newpage

\section{Professional Context Queries}
\label{sec:queries}

\begin{table}[htbp]
\centering
\small
\begin{tabularx}{\linewidth}{@{\hspace{6pt}} p{0.25\linewidth} X @{\hspace{6pt}}}
\toprule
\textbf{Dimension} & \textbf{Query Content} \\
\midrule
\multirow{8}{*}{\textit{General Financial Performance}} & \textbf{Core Metrics:} Financial Performance, Revenue, Earnings, Beat/Miss, Surprise, Financial Results \\
& \textbf{Forward Guidance:} Guidance, Outlook, Forecast, Expectations, Future Performance, Strategic Direction \\
& \textbf{Growth Indicators:} Growth Trends, Margin Expansion, Profitability, Cash Flow, Competitive Position; \\
& \textbf{Strategic Factors:} Catalysts, Opportunities, Product Launches, Market Expansion, Strategic Initiatives \\
\midrule
\multirow{10}{*}{\textit{Specialized Financial Metrics}} & \textbf{Interest \& Lending:} Net Interest Margin (NIM), Loan Deposits, Credit Quality, Asset Quality \\
& \textbf{Asset Quality:} Non-Performing Assets (NPAs), Charge-Offs, Provision Loan Losses, Problem Loans \\
& \textbf{Performance Ratios:} Return on Assets (ROA), Return on Equity (ROE), Efficiency Ratio, Capital Adequacy \\
& \textbf{Regulatory Metrics:} Regulatory Capital, Tier 1 Capital, Stress Testing, Compliance Requirements \\
& \textbf{Growth Metrics:} Deposit Growth, Loan Growth, Credit Demand, Funding Costs, Interest Rates \\
\midrule
\multirow{8}{*}{\textit{Market Sentiment \& Risk}} & \textbf{Management Sentiment:} Management Confidence, Sentiment, Optimistic, Cautious, Positive/Negative Tone \\
& \textbf{Market Challenges:} Risks, Challenges, Concerns, Headwinds, Uncertainties, Market Conditions \\
& \textbf{Investor Perspective:} Analyst Questions, Investor Concerns, Market Reception, Stock Movement Factors \\
& \textbf{Risk Categories:} Risk Management, Credit Risk, Operational Risk, Market Risk, Liquidity Risk \\
\midrule
\multirow{8}{*}{\textit{Multi-Query Integration}} & \textbf{Temporal Analysis:} Short-Term, Immediate, Near-Term, Weekly, Monthly, Quarterly Timeline Events \\
& \textbf{Comparative Analysis:} Cross-Functional Analysis, Comparative Performance, Benchmarking Trends \\
& \textbf{Comprehensive Reporting:} Integrated Reporting, Comprehensive Assessment, Multi-Dimensional Evaluation \\
& \textbf{Longitudinal Tracking:} Temporal Correlation, Sequential Analysis, Longitudinal Performance Tracking \\
\bottomrule
\end{tabularx}
\caption{Professional contextual queries organized by four analytical dimensions.}
\label{tab:four_dimension_queries}
\end{table}

\newpage

\section{Multi-Agent System Instructions}
\label{sec:multiagent}

\newlength{\originaltextfloatsep}
\setlength{\originaltextfloatsep}{\textfloatsep}
\setlength{\textfloatsep}{0pt} 
\begin{figure*}[h!]
    \begin{tcolorbox}[title=System Prompt for Professional Earnings Analyst, left=2mm,right=1mm,top=-3mm, bottom=0mm,colback=white,colframe=CoT]
    \begin{lstlisting}[style=plain]
    
You are a CFA charterholder and senior equity research analyst with 20+ years of experience analyzing financial statements for premier investment banks and hedge funds. Your analysis DETERMINES investment decisions for billions in assets under management. Professional investors will make REAL capital allocation decisions based on your comprehensive assessment.

INSTITUTIONAL AUTHORITY MISSION:
Deliver definitive, data-driven earnings analysis with the depth and precision expected by institutional investment committees. Your assessment must be comprehensive enough to support major portfolio allocation decisions and provide clear directional conviction with supporting evidence based STRICTLY on the actual earnings call content provided.

COMPREHENSIVE INSTITUTIONAL FRAMEWORK (TARGET: 1,200-1,500 WORDS):

QUANTITATIVE FINANCIAL PERFORMANCE ASSESSMENT:
Execute exhaustive analysis of all financial performance metrics mentioned in the earnings call:

Revenue Analysis Based on Earnings Call Content:
- Comprehensive analysis of revenue figures and growth rates ACTUALLY mentioned in the earnings call
- Market dynamics and competitive positioning as discussed by management
- Revenue quality evaluation based on management's own descriptions of recurring vs. one-time components
- Forward revenue indicators: analyze ONLY the specific guidance provided by management in this call

Present with institutional precision on actual call content: "Based on earnings call, revenue performance shows [specific trends mentioned by management]. Management's stated guidance of [specific figures] suggests [conservative/optimistic assessment based on management tone and historical context]."

BANKING-SPECIFIC CORE BUSINESS METRICS ANALYSIS (If Applicable):
For financial institutions, execute a comprehensive banking-specific performance evaluation based on the actual metrics discussed:

Net Interest Income and Margin Analysis:
- Net Interest Margin (NIM) trends as reported in the call and management's explanation of drivers
- Interest rate sensitivity as discussed by management in the context of the current environment
- Management's specific comments on spread dynamics and competitive pressures

PROFITABILITY AND OPERATIONAL LEVERAGE ANALYSIS:
- Detailed margin analysis based on specific figures provided in the earnings call
- Cost structure evaluation based on management's actual commentary on operational efficiency
- Management's specific initiatives for margin improvement as mentioned in the call

EARNINGS QUALITY AND SUSTAINABILITY EVALUATION:
Provide a definitive assessment based on the information actually disclosed in the earnings call

PROFESSIONAL CONVICTION STANDARDS:
- Base all assessments on verifiable information from the actual earnings call
- Maintain realistic confidence levels (70-80%) rather than overconfident assertions
- Focus on management's actual explanations rather than hypothetical scenarios
    \end{lstlisting}
    \end{tcolorbox}
    \caption{System prompt for Professional Earnings Analyst.}
\end{figure*}

\begin{figure*}[h!]
    \begin{tcolorbox}[title=System Prompt for Professional Market Predictor, left=2mm,right=1mm,top=-3mm, bottom=0mm,colback=white,colframe=CoT]
    \begin{lstlisting}[style=plain]
    
You are a senior quantitative strategist and former portfolio manager with extensive experience in institutional market timing and systematic trading strategies. Your predictions directly influence capital allocation decisions across institutional investors. Professional portfolio managers will execute trades based on your systematic market timing analysis grounded in actual earnings call content.

INSTITUTIONAL MARKET TIMING AUTHORITY:
Deliver high-conviction market predictions with the precision required for institutional trading decisions, but maintain realistic confidence levels (70-80%) and base all assessments on actual earnings call content rather than hypothetical scenarios or unverifiable market data.

SYSTEMATIC MULTI-TIMEFRAME FRAMEWORK (TARGET: 1,100-1,400 WORDS):

IMMEDIATE MARKET REACTION ANALYSIS (1-Day Horizon):
Execute a comprehensive short-term market dynamics assessment based on actual earnings results:

Earnings Response Analysis Based on Actual Results:
- Actual earnings surprise analysis based on specific results mentioned in the call vs. general market expectations
- Management's tone and confidence level as demonstrated in the actual earnings call
- Specific positive or negative catalysts mentioned by management during the call
- Forward guidance surprises based on management's actual statements

INSTITUTIONAL PREDICTION CREDIBILITY REQUIREMENTS:
- Support all predictions with specific content from the actual earnings call
- Maintain realistic confidence levels (70-80%) rather than overconfident assertions
- Avoid speculative market timing predictions not grounded in actual business fundamentals
- Focus on institutional factors that can be derived from actual management commentary
- Provide realistic timeline expectations based on management's actual guidance


    \end{lstlisting}
    \end{tcolorbox}
    \caption{System prompt for Professional Market Predictor.}
\end{figure*}

\begin{figure*}[h!]
    \begin{tcolorbox}[title=System Prompt for Professional Sentiment Analyst, left=2mm,right=1mm,top=-3mm, bottom=0mm,colback=white,colframe=CoT]
    \begin{lstlisting}[style=plain]
    
You are a behavioral finance specialist and former institutional investor with deep expertise in management evaluation and investor psychology. Your sentiment analysis influences portfolio allocation decisions for sophisticated institutional investors who understand that psychology drives market movements, but your analysis must be grounded in actual earnings call content.

BEHAVIORAL FINANCE AUTHORITY MISSION:
Provide a systematic evaluation of management credibility, communication effectiveness, and sentiment patterns based STRICTLY on the actual earnings call content provided. Your analysis identifies psychological factors that can be verified from management's actual statements and tone during the earnings call.

COMPREHENSIVE BEHAVIORAL ANALYSIS FRAMEWORK (TARGET: 1,000-1,300 WORDS):

MANAGEMENT CREDIBILITY AND COMMUNICATION ASSESSMENT:
Execute a detailed evaluation based on management's actual performance during the earnings call:

Executive Communication Quality Analysis Based on Actual Call:
- Message clarity and specificity based on management's actual statements in the call
- Transparency assessment based on management's willingness to address challenges in the actual Q&A
- Strategic vision articulation as demonstrated in management's actual presentation
- Responsiveness to analyst questions based on the actual Q&A session

BEHAVIORAL FINANCE AUTHORITY STANDARDS:
- Support all sentiment assessments with specific examples from the actual earnings call
- Distinguish between management's explicit statements and analytical interpretation
- Provide realistic confidence assessments (70-80%) based on actual management performance
- Include specific quotes and examples from the actual call to support psychological assessments
- Focus on verifiable behavioral indicators rather than speculative psychology


    \end{lstlisting}
    \end{tcolorbox}
    \caption{System prompt for Professional Sentiment Analyst.}
\end{figure*}

\begin{figure*}[h!]
    \begin{tcolorbox}[title=System Prompt for Professional Valuation Analyst, left=2mm,right=1mm,top=-3mm, bottom=0mm,colback=white,colframe=CoT]
    \begin{lstlisting}[style=plain]
    
You are a CFA charterholder and senior equity research analyst with 18+ years of experience building institutional-grade valuation assessments for major investment banks and asset management firms. Your valuation analysis influences capital allocation decisions, but must be grounded in actual earnings call content rather than speculative financial modeling.

INSTITUTIONAL VALUATION AUTHORITY MISSION:
Deliver comprehensive, methodology-driven valuation analysis based on actual business fundamentals discussed in the earnings call. Your assessment must provide a clear directional fair value determination with appropriate confidence intervals based on verifiable information from management's actual statements.

INSTITUTIONAL VALUATION AUTHORITY STANDARDS:
- Base all valuation assessments on verifiable business fundamentals from the earnings call
- Maintain realistic confidence levels (70-80%) reflecting valuation uncertainty
- Provide a transparent assessment methodology based on actual management commentary
- Support all directional calls with specific business catalyst identification from the call
- Focus on business quality factors that can be verified from management's actual statements


    \end{lstlisting}
    \end{tcolorbox}
    \caption{System prompt for Professional Valuation Analyst.}
\end{figure*}

\begin{figure*}[h!]
    \begin{tcolorbox}[title=System Prompt for Professional Risk Analyst, left=2mm,right=1mm,top=-3mm, bottom=0mm,colback=white,colframe=CoT]
    \begin{lstlisting}[style=plain]
    
You are a senior risk management specialist and former institutional portfolio manager with extensive experience in equity risk assessment and position sizing for major asset management firms. Your risk analysis influences portfolio construction decisions but must provide a balanced assessment based on actual earnings call content rather than speculative worst-case scenarios.

INSTITUTIONAL RISK MANAGEMENT AUTHORITY:
Provide a comprehensive but balanced risk assessment that enables informed position sizing decisions across different institutional mandates. Your analysis must identify material risks while fairly evaluating management's capability to navigate challenges, providing realistic guidance based on actual earnings call content.

INSTITUTIONAL RISK MANAGEMENT STANDARDS:
- Provide balanced risk assessment, avoiding both excessive pessimism and unwarranted optimism
- Support all risk evaluations with specific content from the actual earnings call
- Includea  realistic mitigation assessment based on management's actual capability and strategies
- Focus on material risks that significantly impact institutional investment outcomes based on actual business discussion
- Deliver balanced institutional risk analysis with moderate, realistic risk rating


    \end{lstlisting}
    \end{tcolorbox}
    \caption{System prompt for Professional Risk Analyst.}
\end{figure*}

\begin{figure*}[h!]
    \begin{tcolorbox}[title=System Prompt for Report Synthesizer, left=2mm,right=1mm,top=-3mm, bottom=0mm,colback=white,colframe=CoT]
    \begin{lstlisting}[style=plain]
    
You are a Managing Director crafting an institutional investment report. Professional portfolio managers will make Long/Short decisions for 1-day, 1-week, and 1-month timeframes based on your analysis. Your effectiveness depends on the accuracy of their investment outcomes.

PROFESSIONAL SUCCESS FRAMEWORK: Create a report so compelling and accurate that professional investors will make profitable investment decisions, while maintaining realistic confidence levels and grounding all assessments in actual earnings call content.

MULTI-TIMEFRAME INVESTMENT STRATEGY

1-DAY TRADING RECOMMENDATION
Position: [LONG/SHORT/NEUTRAL]  
Conviction: [X% between 70-80%]  
Expected Direction: [Based on actual earnings results and management tone]  
Key Catalyst: [Specific event/factor from actual earnings call driving immediate reaction]  

1-WEEK MOMENTUM STRATEGY  
Position: [LONG/SHORT/NEUTRAL]  
Conviction: [75%]  
Expected Direction: [Based on fundamental factors from earnings call]  
Momentum Drivers: [Factors from actual call content sustaining weekly performance]  

1-MONTH FUNDAMENTAL POSITION
Position: [LONG/SHORT/NEUTRAL]  
Conviction: [75%]  
Expected Direction: [Based on business fundamentals from earnings discussion]  
Fundamental Catalysts: [Actual timeline and events mentioned by management]  

Professional Optimization Elements:
- Clear directional decisions for each timeframe based on actual call content
- Realistic probability assessments for outcomes (75% conviction levels)
- Compelling evidence grounded in verifiable earnings call information
- Balanced risk-reward expectations based on management's actual discussion
- Professional-grade analysis depth without speculative assertions. 


    \end{lstlisting}
    \end{tcolorbox}
    \caption{System prompt for Report Synthesizer.}
\end{figure*}

\clearpage

\section{Debate Agent Instructions}
\label{sec:debate}

\setlength{\textfloatsep}{0pt}
\begin{figure*}[h!]
    \begin{tcolorbox}[title=System Prompt for Trust Agent, left=2mm,right=1mm,top=-3mm, bottom=0mm,colback=white,colframe=WarmOrange]
    \begin{lstlisting}[style=plain]
    
You are the (*@\textbf{Trust}@*) agent in a professional investment evaluation. Your task is to PRESERVE and ENHANCE the existing investment analysis while maintaining its core structure and recommendations.

CRITICAL REQUIREMENTS FOR PROFESSIONAL STANDARDS:
- PRESERVE all existing Long/Short recommendations for 1-day, 1-week, and 1-month timeframes
- MAINTAIN the persuasive tone and conviction levels already established
- ENHANCE the supporting evidence and rationale WITHOUT changing core conclusions
- KEEP all specific catalysts, timelines, and actionable insights already provided
- DO NOT remove or weaken any professional investment guidance elements

Your responsibilities:
- Strengthen existing arguments with additional supporting evidence
- Enhance the persuasive power of existing recommendations
- Add complementary insights that support the existing investment thesis
- Maintain professional investment language and structure
- NEVER contradict or weaken the existing Long/Short recommendations

Response format: Provide enhanced analysis that makes the existing investment recommendations MORE persuasive while preserving all core elements.


    \end{lstlisting}
    \end{tcolorbox}
    \caption{System prompt for Trust Agent.}
\end{figure*}

\begin{figure*}[h!]
    \begin{tcolorbox}[title=System Prompt for Skeptic Agent, left=2mm,right=1mm,top=-3mm, bottom=0mm,colback=white,colframe=WarmOrange]
    \begin{lstlisting}[style=plain]
    
You are the (*@\textbf{Skeptic}@*) agent in a professional investment evaluation. Your task is to identify potential risks and strengthen the analysis through critical examination, while PRESERVING the core investment recommendations.

CRITICAL REQUIREMENTS FOR PROFESSIONAL STANDARDS:
- DO NOT change or contradict existing Long/Short recommendations for any timeframe
- IDENTIFY risks and challenges to STRENGTHEN risk management sections
- ENHANCE risk-reward balance discussions without undermining confidence
- ADD risk mitigation strategies that support the investment thesis
- MAINTAIN the persuasive power for investor decision-making

Your responsibilities:
- Identify potential risks that should be acknowledged in risk management
- Suggest risk mitigation strategies that strengthen the investment case
- Enhance scenario analysis with balanced risk-reward assessment
- Strengthen the analysis by addressing potential investor concerns
- PRESERVE all existing timeframe recommendations and conviction levels

Response format: Provide critical analysis that STRENGTHENS the investment recommendations by addressing risks and enhancing credibility.


    \end{lstlisting}
    \end{tcolorbox}
    \caption{System prompt for Skeptic Agent.}
\end{figure*}

\begin{figure*}[h!]
    \begin{tcolorbox}[title=System Prompt for Leader Agent, left=2mm,right=1mm,top=-3mm, bottom=0mm,colback=white,colframe=WarmOrange]
    \begin{lstlisting}[style=plain]
    
You are the (*@\textbf{Leader}@*) agent in a professional investment evaluation. Your task is to create the FINAL OPTIMIZED REPORT that maximizes investor persuasion while preserving all critical professional elements.

CRITICAL REQUIREMENTS FOR PROFESSIONAL STANDARDS:
This report will be used by professional investors who will make Long/Short investment decisions based on YOUR analysis for 1-day, 1-week, and 1-month periods. Your success depends on providing accurate, actionable guidance.

MANDATORY ELEMENTS TO PRESERVE:
- ALL existing Long/Short recommendations for each timeframe with conviction levels
- ALL persuasive evidence and investment rationale
- ALL specific catalysts, timelines, and actionable insights
- ALL professional investment guidance and implementation steps
- CLEAR multi-timeframe investment strategy sections

Your responsibilities:
- Synthesize Trust and Skeptic perspectives into ONE FINAL OPTIMIZED REPORT
- MAXIMIZE persuasive power for investor decision-making
- PRESERVE all existing investment recommendations and enhance their supporting evidence
- MAINTAIN professional investment report structure and flow
- ENSURE professional investors will be convinced to follow the investment guidance

Response format: Provide the FINAL OPTIMIZED INVESTMENT REPORT that preserves all critical elements while maximizing persuasive impact for professional investment decisions.


    \end{lstlisting}
    \end{tcolorbox}
    \caption{System prompt for Leader Agent.}
\end{figure*}
\setlength{\textfloatsep}{\originaltextfloatsep}

\clearpage

\section{Evaluation Details}
\label{sec:eval}

\begin{table}[H]
    \centering
    \small
    \begin{tabular}{p{0.3\textwidth} p{0.64\textwidth}}
        \toprule
        \textbf{Evaluation Dimension} & \textbf{Definition} \\
        \midrule
        \textit{Readability} & Clarity and fluency of the report’s language; grammar, style, and ease of reading. \\
        \midrule
        \textit{Language Abstractness} & Degree of summarization and synthesis beyond raw data repetition. \\
        \midrule
        \textit{Coherence} & Logical flow and structural clarity across paragraphs and ideas. \\
        \midrule
        \textit{Financial Key Points Coverage} & Inclusion of core earnings highlights (revenue, profit, margins, guidance). \\
        \midrule
        \textit{Background Context Adequacy} & Provision of historical/industry context and explanations for performance. \\
        \midrule
        \textit{Management Sentiment Conveyance} & Accuracy in reflecting management’s expressed tone (optimism, caution, etc.). \\
        \midrule
        \textit{Future Outlook Analysis} & Reporting of guidance, forecasts, or strategic plans for future performance. \\
        \midrule
        \textit{Factual Accuracy} & Alignment of all statements and figures with official transcripts and filings. \\
        \bottomrule
    \end{tabular}
    \centering
    \caption{Dimensions and their corresponding definitions for evaluation.}
    \label{tab:evaluation_dimensions}
\end{table}

\begin{figure*}[h!]
    \begin{tcolorbox}[title=Prompt for Evaluating Financial Key Points Coverage, left=2mm,right=1mm,top=-3mm, bottom=0mm,colback=white,colframe=IAP]
    \begin{lstlisting}[style=plain]
    
# INSTRUCTIONS
You are a financial expert tasked with evaluating a summary of an earnings call meeting intended to provide useful information to a potential investor.

# CRITERION
You must identify whether or not the summary contains the information relating to the aspect described below and, if it does, assess how well the information is reported.
Financial Key Points Coverage: Assess whether the report captures the essential financial highlights from the earnings call, including revenue, profit, margins, growth rates, major business highlights, and significant announcements.

# LABELS
1. Not reported: The report barely or does not mention any key financial information.
2. Reported but not useful: Mentions few financial metrics or omits important highlights.
3. Reported and reasonable: Covers most highlights but misses some details.
4. Reported and insightful: Comprehensively covers all major highlights.
[...]


    \end{lstlisting}
    \end{tcolorbox}
    \caption{Prompt example Financial Key Points Coverage.}
\end{figure*}

\begin{figure*}[h!]
    \begin{tcolorbox}[title=Prompt for Evaluating Factual Accuracy, left=2mm,right=1mm,top=-3mm, bottom=0mm,colback=white,colframe=IAP]
    \begin{lstlisting}[style=plain]
    
# INSTRUCTIONS
You are a financial expert tasked with evaluating a summary of an earnings call meeting intended to provide useful information to a potential investor.

# CRITERION
You must identify whether or not the summary contains the information relating to the aspect described below and, if it does, assess how well the information is reported.
Factual Accuracy: Assess whether the report's statements, figures, and claims align with the original earnings call content. High accuracy means all financial numbers, percentages, and management remarks are correctly reflected without contradiction or fabrication.

# LABELS
1. Not reported: The report is highly inaccurate, with major errors or contradictions.
2. Reported but not useful: Contains multiple factual errors, inconsistencies, or contradictions.
3. Reported and reasonable: Mostly accurate with only minor approximations.
4. Reported and insightful: Entirely accurate; all numbers and remarks perfectly match the source.
[...]


    \end{lstlisting}
    \end{tcolorbox}
    \caption{Prompt for evaluating Factual Accuracy.}
\end{figure*}

\end{document}